\documentclass[11pt]{article}
\usepackage{acl}
\usepackage{times}
\usepackage{latexsym}
\usepackage[T1]{fontenc}
\usepackage[utf8]{inputenc}
\usepackage{microtype}
\usepackage{inconsolata}
\usepackage{booktabs}
\usepackage{array}
\usepackage{graphicx}
\usepackage{amsmath}
\usepackage{amssymb}
\usepackage{multirow}
\usepackage{enumitem}

\newcommand{\modelname}[1]{\textit{#1}}
\newcommand{\true}{\texttt{true}}
\newcommand{\false}{\texttt{false}}
\newcommand{\unknown}{\texttt{unknown}}
\newcolumntype{L}[1]{>{\raggedright\arraybackslash}p{#1}}

\newcommand{\inputtable}[1]{%
  \edef\tablepath{tables/#1.tex}%
  \expandafter\input\expandafter{\tablepath}%
}

\newcommand{\appendixwidetable}[4]{%
\begin{table*}[!t]
\centering
\small
\resizebox{#1\textwidth}{!}{%
\inputtable{#2}%
}
\captionsetup{hypcap=false}
\caption{#3}
\label{#4}
\end{table*}
}
\newcommand{\appendixcolumntable}[4]{%
\begin{table}[!t]
\centering
\small
\resizebox{#1\columnwidth}{!}{%
\inputtable{#2}%
}
\captionsetup{hypcap=false}
\caption{#3}
\label{#4}
\end{table}
}
\newcommand{\appendixinlinecolumntable}[4]{%
\par\noindent
\begin{minipage}{#1\columnwidth}
\centering
\small
\resizebox{\linewidth}{!}{%
\inputtable{#2}%
}\par
\captionsetup{type=table,hypcap=false}
\caption{#3}
\label{#4}
\end{minipage}
\par
}
\newcommand{\appendixstackedtwotables}[8]{%
\begin{table*}[!t]
\centering
\small
\resizebox{#1\textwidth}{!}{%
\inputtable{#2}%
}
\captionsetup{hypcap=false}
\caption{#3}
\label{#4}
\par
\resizebox{#5\textwidth}{!}{%
\inputtable{#6}%
}
\captionsetup{hypcap=false}
\caption{#7}
\label{#8}
\end{table*}
}
\newcommand{\appendixsidebyside}[8]{%
\begin{table*}[!t]
\centering
\small
\begin{minipage}[t]{0.48\textwidth}
\centering
\resizebox{#1\textwidth}{!}{%
\inputtable{#2}%
}
\captionsetup{hypcap=false}
\captionof{table}{#3}
\label{#4}
\end{minipage}\hfill
\begin{minipage}[t]{0.48\textwidth}
\centering
\resizebox{#5\textwidth}{!}{%
\inputtable{#6}%
}
\captionsetup{hypcap=false}
\captionof{table}{#7}
\label{#8}
\end{minipage}
\end{table*}
}

\title{Wrong Prediction, Right Answer: Recovering Evidence from Collapsed LLM Sequence Scores}

\author{
  \textbf{Qiyao Yan\textsuperscript{1,2}} \and
  \textbf{Chenpeng Wang\textsuperscript{3}} \and
  \textbf{Liangming Pan\textsuperscript{1,3,4}\thanks{Corresponding Author.}} \\
  \textsuperscript{1}State Key Laboratory of Multimedia Information Processing, Peking University\\
  \textsuperscript{2}School of Computer Science, Peking University\\
  \textsuperscript{3}YiXin-AILab, YIXIN, Beijing, China\\
  \textsuperscript{4}Beijing Academy of Artificial Intelligence, Beijing, China\\
  \texttt{yqy.1@foxmail.com}, \texttt{liangmingpan@pku.edu.cn}
}

\begin{document}
\maketitle

\begin{abstract}
When a large language model fails a reasoning task, it is often assumed to lack the underlying capability. However, this conflates a genuine absence of reasoning with a late-stage output bottleneck. We observe a consistent readout gap across diverse reasoning benchmarks: hidden-state probes successfully decode correct answers even when native sequence scoring completely collapses due to structural biases. To test whether instance-specific logic survives this collapse, we introduce a diagnostic protocol using a minimal, target-label-free additive correction. Fitting just two parameters on as few as 25 unlabeled examples recovers $9$--$34$ accuracy points for \modelname{Qwen3.5} models, transferring successfully to \modelname{OLMo-2-1B} and \modelname{Llama-3.1-8B}. Crucially, these recovered decisions persist on hard instances unresolved by simple lexical overlap and significantly exceed count-preserving permutation baselines. Our results definitively show that many apparent zero-shot reasoning deficits are merely expression failures masking intact internal logic, urging a narrower interpretation of benchmark evaluations.
\end{abstract}

\section{Introduction}

As large language models (LLMs) continue to advance \cite{openai2023gpt4,dubey2024llama,anthropic2024claude}, standard evaluation relies on a straightforward assumption: if a model predicts the wrong final answer, it lacks the underlying reasoning capability. However, this view mixes up two fundamentally distinct mechanisms: a model's internal understanding and its outward expression. A model might successfully figure out the correct answer behind the scenes, yet fail to express it through its final generated token. Therefore, looking only at the final output observed by an evaluator can be highly misleading.

This disconnect between internal knowledge and outward expression has been revealed by a series of studies. Investigations into model internals (often using hidden-state probes) demonstrate that LLMs actually compute the correct reasoning steps and answers deep inside their layers, well before reaching the final output stage \cite{alain2016understanding,belrose2023tuned,burns2023discovering,bao-etal-2025-probing,servedio-etal-2025-hidden,maiya-etal-2025-improving,sun-etal-2025-probing,yuchi-etal-2026-llms}. The problem typically occurs at the very end of the process. When the model translates its internal representations into actual vocabulary words, various structural biases kick in. For example, the model might have a built-in preference for frequent labels like \unknown~or be thrown off by the phrasing of the prompt \cite{zhao2021calibrate,holtzman2021surface,xia-etal-2025-influences,li-etal-2025-large-language-models,huang-etal-2026-investigating}. As a result, the specific correct answer for a given question gets overwritten by these general biases, causing the final prediction to collapse into what looks like a random guess.

Existing works try to tackle this gap from two different directions, but both fall short in crucial ways. On one hand, while hidden-state probes prove that the model can represent the answer, they act as flexible external observers; they do not prove that the model's native generation process actually uses this information \cite{hewitt2019designing,voita2020information,pimentel2020information,belinkov2022probing}. On the other hand, output calibration methods attempt to fix the final predictions by mathematically rebalancing the output probabilities to mitigate biases \cite{zhao2021calibrate,holtzman2021surface,sanz-guerrero-von-der-wense-2025-mitigating,li-etal-2025-calibraeval,wang-liu-2025-beyond,xiao2025restoring,nakkiran2026trained}. While these techniques improve overall accuracy, this aggregate success can be highly deceptive. Simply shifting the global distribution of answers (such as forcing the model to output \true~and \false~at equal rates) does not guarantee that the model's underlying logical reasoning for individual questions was actually fixed or utilized.

To address this ambiguity, we introduce a comprehensive diagnostic protocol to systematically trace exactly where the model's reasoning signal gets lost. We first compare the model's hidden-state probes, single-token logits, and full generative sequence scores to pinpoint the exact stage of the prediction collapse. Once we isolate this late-stage scoring bottleneck, we test whether the model's native sequence scores still hide correct, example-specific logic. We do this by applying a \textit{minimal, target-label-free additive correction} to the raw scores. Because this intervention is mathematically restricted---merely shifting the global decision boundaries using only two parameters without accessing any ground-truth labels---it cannot teach the model new reasoning skills or alter how it internally ranks candidate answers for a specific prompt. Therefore, if the model's accuracy dramatically recovers after this minor adjustment, it provides \textit{definitive proof} that the correct logical alignment was preserved in the native outputs all along, and the apparent failure was merely a \textit{late-stage scoring distortion}.

We implement this protocol across a deliberate progression of reasoning tasks, starting from controlled symbolic deduction to natural-language benchmarks like ProofWriter, ANLI, and FOLIO. These tasks provide clear ground truths and balanced labels, making them ideal for isolating logical deduction from general text generation. Testing on the \modelname{Qwen3.5} model family \cite{qwen35}, we find a consistent gap: probes easily decode the correct answer, while normal output scoring fails completely. However, by applying our simple two-parameter correction using as few as 25 unlabeled examples, we recover $9$--$34$ accuracy points. Importantly, this recovery holds up against strict stress tests, proving it relies on \textit{genuine semantic reasoning} rather than shallow word-matching shortcuts. The effect also transfers successfully to \modelname{OLMo-2-1B} \cite{groeneveld2024olmo} and \modelname{Llama-3.1-8B} \cite{dubey2024llama}.

\begin{figure*}[t]
\centering
\includegraphics[width=\textwidth]{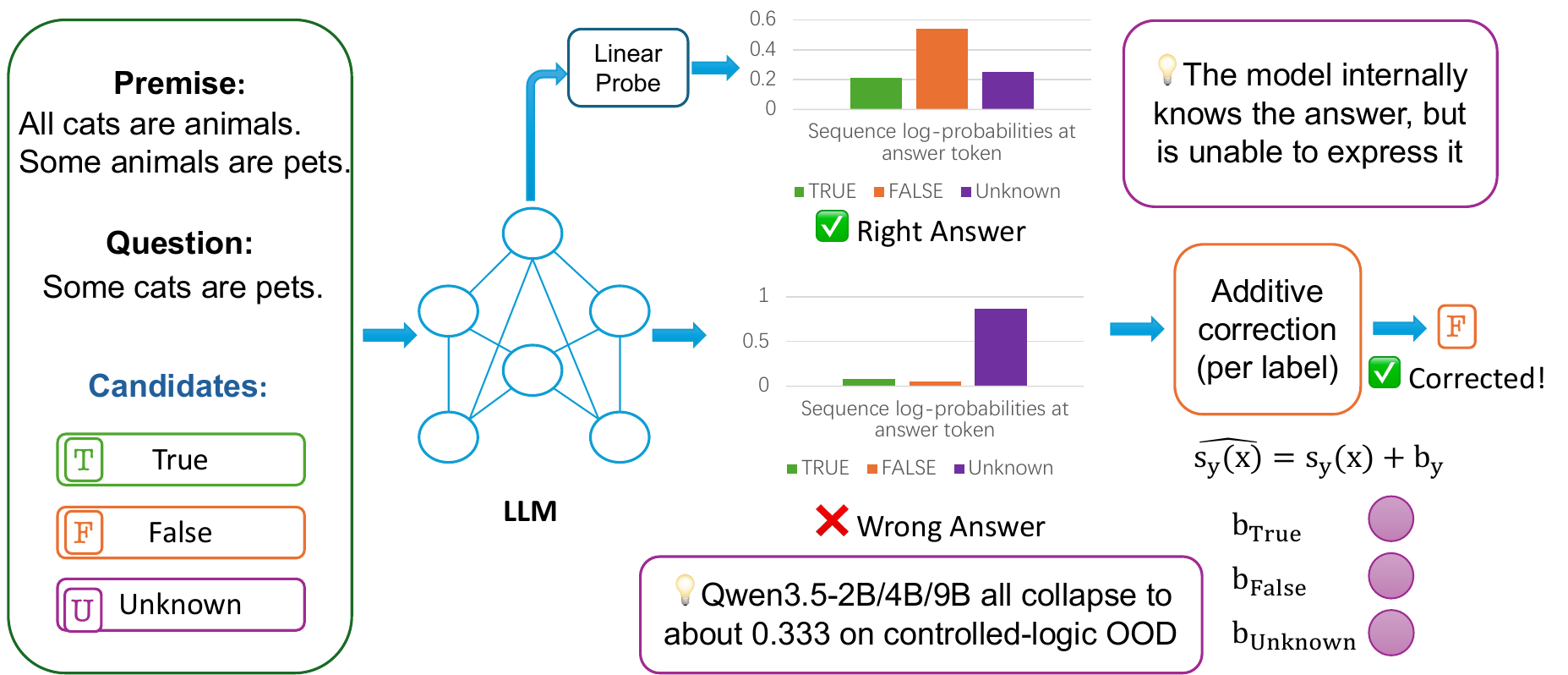}
\caption{
\textbf{Diagnosing collapsed candidate-answer scores.} Raw candidate-answer
string scores can select the wrong label even when a linear probe decodes the
gold label from a hidden state. We apply a prior-conditioned additive offset
($s_y(x) \mapsto s_y(x)+b_y$) as a diagnostic intervention; held-out rescue and
permutation controls test whether the correction exposes example-specific
structure.}
\label{fig:protocol}
\end{figure*}

\section{Related Work}

\paragraph{Probing versus output scoring.}
Research into model internal representations often relies on linear probes trained on frozen hidden states. These studies show that models encode substantial task-relevant information---including truth directions, factuality, arithmetic steps, and chain-of-thought answer formation---well before the final output token \cite{alain2016understanding,nostalgebraist2020logit,belrose2023tuned,bao-etal-2025-probing,servedio-etal-2025-hidden,maiya-etal-2025-improving,sun-etal-2025-probing,yuchi-etal-2026-llms,kudo-etal-2026-llms}. Probe accuracy, however, bypasses the model's own candidate-answer scoring and can arise from statistical features that the model's native prediction path ignores \cite{hewitt2019designing,voita2020information,pimentel2020information,belinkov2022probing}. Probes are flexible external observers: their success demonstrates theoretical capacity, not practical utilization. As \citet{burns2023discovering} show, models can represent the correct answer internally while still outputting the wrong token. The challenge, then, is to identify what blocks the correct answer from reaching the output.

\paragraph{Softmax bottlenecks and unembedding misalignment.}
One explanation for this obstruction lies in the geometry of the language modeling head. The softmax bottleneck \cite{yang2018softmax} shows that the final linear projection and softmax constrain the expressiveness of the output layer: even when a complex reasoning manifold exists in the final hidden state, mapping it into the vocabulary space through a single static linear transformation (the unembedding matrix) can distort it. Post-training can amplify these biases: instruction tuning and reinforcement learning from human feedback shift the baseline probabilities of structural tokens \cite{xiao2025restoring}, which can push native scoring toward safe or frequent labels (e.g., \unknown~or \texttt{neutral}) irrespective of the context vector.

\paragraph{Output bias and per-example recovery.}
Language model predictions over fixed choices are susceptible to systematic distortions, including label frequency \cite{zhao2021calibrate}, surface-form competition \cite{holtzman2021surface}, and prompt wording \cite{xia-etal-2025-influences,li-etal-2025-large-language-models,huang-etal-2026-investigating}. While post-hoc corrections---such as verbal normalization and diagonal scaling \cite{wang-liu-2025-beyond,sanz-guerrero-von-der-wense-2025-mitigating,xiao2025restoring,li-etal-2025-calibraeval,nakkiran2026trained}---can mitigate these biases, their success is measured almost exclusively by aggregate accuracy or expected calibration error. These metrics are limited as diagnostic tools because they cannot distinguish a correction that fixes individual reasoning paths from one that only reshapes the global label distribution. We therefore pair additive offsets with per-example gain counts and label-count permutation baselines, which verify that corrected decisions resolve the right individual instances.

\paragraph{Logical-reasoning benchmarks as testbeds.}
Controlled deduction tasks \cite{clark2020transformers,saparov2022greedy} and natural-language benchmarks such as ProofWriter \cite{tafjord2021proofwriter}, FOLIO \cite{han2022folio}, and ANLI \cite{nie2019adversarial} provide known ground truth, balanced labels, and controllable proof depth, making them well suited for auditing individual components of a prediction pipeline. Instead of treating these benchmarks as leaderboards where a wrong answer is equated with an absence of reasoning capacity, we use their structural controls to disentangle representation failures from scoring bottlenecks.

\section{Method}
\label{sec:method}

To localize where a final-answer error originates, we use a three-stage diagnostic pipeline (Figure~\ref{fig:protocol}): (i)~establish whether the answer is encoded in hidden states; (ii)~test whether the frozen output layer exposes this information; and (iii)~apply a minimal, prior-conditioned additive correction without target labels, verifying any recovery at the per-example level. No training objective is added to the model itself; every component operates on frozen outputs.

\paragraph{Task and full candidate-answer string score.}
Each example consists of a prompt $x$ and a label $y$ drawn from a fixed set (e.g., \{\true,\false,\unknown\}). The prompt ends with a literal answer slot, \texttt{Answer: <answer>}. We evaluate candidates by scoring fixed continuations, such as \texttt{true</answer>}. The \emph{full candidate-answer string score} is the sum of teacher-forced log-probabilities over this complete continuation:
$$
    s_y(x) = \sum_{t=1}^{|a_y|}\log p_\theta(a_{y,t}\mid x,r,a_{y,<t}),
$$
\looseness=-1 where $r$ is the prompt suffix and $a_y$ is the fixed string. The raw prediction is $\arg\max_y s_y(x)$. This step adds no auxiliary task heads or parameters, so the scores reflect native zero-shot behavior.

\paragraph{Probe training.}
All probes are multinomial logistic regressions fitted on frozen hidden states from in-domain examples and evaluated on disjoint held-out splits; no gradient reaches the model. Training sizes, seeds, and padding follow the format-matched protocol listed in Appendix~\ref{app:data} (e.g., 2,000 training rows and three fixed seeds for the repaired 4B runs); training-size sensitivity appears in Appendix~\ref{app:readout}.

\paragraph{Mathematical formulation of the scoring gap.}
To separate linearly decodable information from native output representations, we use position-matched readouts. Let $h^a(x) \in \mathbb{R}^d$ be the hidden state immediately following the \texttt{Answer: <answer>} tag, before any label token is generated. At this position, the \emph{same-position label logits} are computed via the frozen unembedding matrix $W \in \mathbb{R}^{V \times d}$ and bias vector $b \in \mathbb{R}^V$. For a candidate token $y$, the native score is determined by the pre-trained geometry alone: $z_y = h^a(x)^\top w_y + b_y$. If the structural bias $b_y$ is large, or if the norm $\|w_y\|$ favors a specific vocabulary subset, the prediction collapses regardless of the logic encoded in $h^a(x)$.

An \emph{answer-slot probe}, by contrast, fits an independent separating hyperplane $\theta_y$ via logistic regression, outside the vocabulary-projection constraints. Probing at this shared state $h^a(x)$ avoids cascading effects from generated tokens and isolates the context vector available to the model at the decision point; a \emph{prompt-end probe} applies the same readout to the final state of the original prompt, before the answer tag is appended.

\paragraph{Prior-conditioned offset correction.}
Building on established additive calibration techniques \cite{zhao2021calibrate}, we use a low-capacity decision offset as our diagnostic intervention to counteract the $b_y$ dominance described above. Given an unlabeled in-domain score set $D_{\mathrm{id}}$, the corrected prediction is:
$$
    \hat y_c(x) = \arg\max_y \; s_y(x) + c_y.
$$
We fix one reference offset to zero (a uniform shift preserves the argmax), leaving two free parameters for three labels. Given a target label prior $\pi$ and the induced prediction fraction $q_c(y;D_{\mathrm{id}})$, we choose the offsets by deterministic grid search to minimize the squared difference:
$$
    c^\star = \arg\min_c \sum_y \left(q_c(y;D_{\mathrm{id}})-\pi_y\right)^2 .
$$
This formulation is deliberately constrained: with only two degrees of freedom for the entire dataset, the intervention cannot learn a new reasoning-task mapping. Because the offsets are global, they cannot alter within-label rankings; the correction acts only as a threshold shift.

To make this concrete, consider a \modelname{Qwen3.5-9B} lexical-OOD example (gold label \true) whose raw scores collapse to \unknown. Offsets fitted on an unlabeled ID set rebalance the candidates so that the correct one wins:
\begin{center}
\small
\resizebox{0.95\columnwidth}{!}{%
\begin{tabular}{lrrr}
\toprule
Candidate continuation & Raw score & Offset & Corrected score \\
\midrule
\texttt{false</answer>} & $-6.18$ & $+1.55$ & $-4.63$ \\
\texttt{true</answer>} & $-5.89$ & $+3.25$ & $-2.64$ \\
\texttt{unknown</answer>} & $-3.49$ & $0.00$ & $-3.49$ \\
\bottomrule
\end{tabular}
}
\end{center}

\paragraph{Evaluation protocol and controls.}
The offset-fit set is always disjoint from the evaluation set. We consider a recovery valid only if it yields positive per-example gains \emph{and} survives three controls:
\begin{enumerate}[leftmargin=*,nosep]
    \item \textbf{TF-IDF-missed slice:} Language models can exploit spurious correlations---such as shallow word overlap between premise and hypothesis---to score well without reasoning (the ``Clever Hans'' effect). Restricting evaluation to examples that a bag-of-words classifier fails on removes instances solvable by surface matching alone.
    \item \textbf{Label-count permutation baselines:} Accuracy is inflated if a correction only matches the true label distribution of a balanced dataset. We randomly permute the corrected predictions while preserving their label multiset; the resulting \emph{permutation gap} quantifies per-example alignment beyond label counts.
    \item \textbf{Deterministic subsampling:} Fitting offsets on very small subsets (25 to 500 examples) tests whether the correction extracts a pre-existing threshold shift, ruling out data-intensive relearning of a task mapping.
\end{enumerate}

\section{Experiments}

\subsection{Setup and design rationale}

Our evaluation scales the complexity of the reasoning environment in four steps. We begin with \textbf{controlled logic}, which isolates multi-hop deductive rules and strips away world knowledge to test pure algorithmic syntax. \textbf{ProofWriter} embeds the same style of logic in natural language, adding linguistic variance. \textbf{ANLI} tests whether the findings survive human-written adversarial noise, and \textbf{FOLIO} probes zero-shot transfer to formal first-order logic. This progression ensures that an observed scoring collapse is not an artifact of a single linguistic domain.

All four tasks are cast into the same forced-choice format with a literal answer slot. For controlled logic, the ID fit split and each OOD split contain 1,000 examples, generated with fixed scripts and seeds and balanced over the three labels; distractor rules are label-balanced, so shallow matching does not solve the task. For ProofWriter, offsets are fitted on 1,000 in-domain score rows and evaluated on a disjoint 1,000-example OOD split. ANLI uses R1 as the in-domain fit split and the adversarial R2 as the held-out split. FOLIO serves as a sparsity diagnostic: offsets are fitted on the 203 validation examples and evaluated on 1,001 prepared training examples. Full construction details appear in Appendix~\ref{app:data}.

Our design holds the model family fixed, so that comparisons across scoring methods are not confounded by architectural differences: the central evaluation covers a protocol-matched, post-trained \modelname{Qwen3.5-2B/4B/9B} sweep, while cross-family models (\modelname{OLMo-2-1B} \cite{groeneveld2024olmo}, \modelname{Llama-3.1-8B} \cite{dubey2024llama}, and \modelname{Pythia} \cite{biderman2023pythia}) are tested only on ProofWriter, where they probe the generalization boundary. Appendix~\ref{app:data} records the protocol details and model revisions.

\subsection{Readout gap: string scoring fails while answer-slot evidence remains}
\label{sec:readout-gap}

To isolate candidate-answer likelihood as the failing step, we first test whether task-relevant evidence survives in hidden states when sequence scoring collapses. If the answer is absent, probes should fail alongside native scoring.

Table~\ref{tab:qwen_posttrained_position_matched} confirms this divergence. For post-trained \modelname{Qwen3.5-2B}, answer-slot probe accuracies reach $0.918$ (in-domain) and $0.661$ (lexical OOD), while the full candidate-answer string score stays at chance ($0.333$) on every split. The gap widens for \modelname{Qwen3.5-9B}, whose answer-slot probes reach $0.972$ (ID) and $0.830$ (lexical OOD) against the same collapsed string scores. The same-position label logits lag well behind the probes, confirming that the hidden state encodes the correct answer even where the frozen output matrix does not expose it.

The mismatch is thus geometric rather than representational: the answer is decodable from the hidden state, but the pre-trained unembedding geometry is not aligned with the direction that encodes it. Wrong sequence predictions therefore cannot be read as an absence of reasoning.

\looseness=-1 Nor is the divergence a post-training artifact. Table~\ref{tab:qwen_base_position_matched} repeats the comparison on the \modelname{Qwen3.5} Base checkpoints, which never underwent instruction tuning: \modelname{Qwen3.5-4B-Base} answer-slot probes reach $0.968$ (ID) and $0.751$ (lexical OOD) while the string score stays at chance. The base rows also localize the loss more finely. For \modelname{Qwen3.5-9B-Base}, same-position label logits reach $0.627$ (ID) and $0.607$ (lexical OOD), yet the full string score remains at $0.374$/$0.361$: even where the vocabulary head partially exposes the answer at the decision position, accumulating log-probability over the candidate string destroys the margin again. The bottleneck therefore sits in sequence-level aggregation as much as in the unembedding geometry.

Three further audits bound alternative accounts (Appendix~\ref{app:readout}). First, the readout gap is statistically solid: on \modelname{Qwen3.5-9B} the answer-slot probe exceeds the string score by $+0.488$ (95\% CI $[+0.452, +0.523]$, paired $p<.001$), and the same-position label logits exceed the string score by $+0.237$ $[+0.180, +0.293]$, so information is lost at the vocabulary projection \emph{and} again at sequence aggregation. Second, probes fitted to randomized labels stay at chance ($0.304$--$0.341$), so probe success reflects answer-related structure rather than arbitrary decodability. Third, the collapse is invariant to scoring convention: mean-normalized scores reproduce the summed-score collapse exactly (e.g., \modelname{Qwen3.5-9B} predicts \unknown~for 99.9\% of lexical-OOD examples under both), ruling out length normalization as the cause.

\begin{table*}[t]
\centering
\small
\resizebox{\textwidth}{!}{%
% Auto-generated from phase1_probe_summary.csv, position_matched_readout_summary.csv,
% and protocol-matched qwen4b_answer_slot_repair.csv when available.
% Regenerate with: python scripts/render_paper_tables.py --table qwen_posttrained_position_matched
\begin{tabular}{llccccc}
\toprule
Model & Split & Prompt-end probe & Answer-slot probe & Same-position label logits & First-label-token score & Full string score \\
\midrule
\textit{Qwen3.5-2B} & \texttt{id} & 0.908 & 0.918 & 0.333 & 0.333 & 0.333 \\
\textit{Qwen3.5-2B} & \texttt{depth} & 0.580 & 0.583 & 0.333 & 0.333 & 0.333 \\
\textit{Qwen3.5-2B} & \texttt{lexical} & 0.516 & 0.661 & 0.333 & 0.333 & 0.333 \\
\midrule
\textit{Qwen3.5-4B} & \texttt{id} & 0.951 & 0.962 & 0.575 & 0.572 & 0.333 \\
\textit{Qwen3.5-4B} & \texttt{depth} & 0.761 & 0.858 & 0.521 & 0.515 & 0.333 \\
\textit{Qwen3.5-4B} & \texttt{lexical} & 0.644 & 0.774 & 0.561 & 0.552 & 0.333 \\
\midrule
\textit{Qwen3.5-9B} & \texttt{id} & 0.975 & 0.972 & 0.573 & 0.572 & 0.333 \\
\textit{Qwen3.5-9B} & \texttt{depth} & 0.799 & 0.821 & 0.480 & 0.478 & 0.333 \\
\textit{Qwen3.5-9B} & \texttt{lexical} & 0.611 & 0.830 & 0.574 & 0.568 & 0.333 \\
\bottomrule
\end{tabular}%
}
\caption{\textbf{Position-matched post-trained \modelname{Qwen3.5} scoring comparison.}
The prompt-end probe reads the final state of the original prompt; the
answer-slot probe reads the state after literal \texttt{Answer: <answer>} and
before any label; same-position label logits apply the frozen output matrix at
that exact state; and the full candidate-answer string score sums
teacher-forced log-probabilities over the fixed continuation. All columns use
the same 1,000 examples per split. The repaired 4B probe values use the
format-matched 2,000-example probe training run (three fixed seeds).}
\label{tab:qwen_posttrained_position_matched}
\end{table*}

\subsection{Label-bias correction recovers per-example evidence}
\label{sec:sequence-recovery}

\looseness=-1 A pure label-frequency bias distorts the output distribution while leaving example-level rankings intact. We test whether sequence scoring suffers from this specific bottleneck by applying the two-parameter offset correction and tracking per-example macro-F1: if native scoring retains task-relevant structure, the correction should produce positive paired gains, not merely a balanced histogram.

\begin{figure}[t]
\centering
\includegraphics[width=0.76\columnwidth]{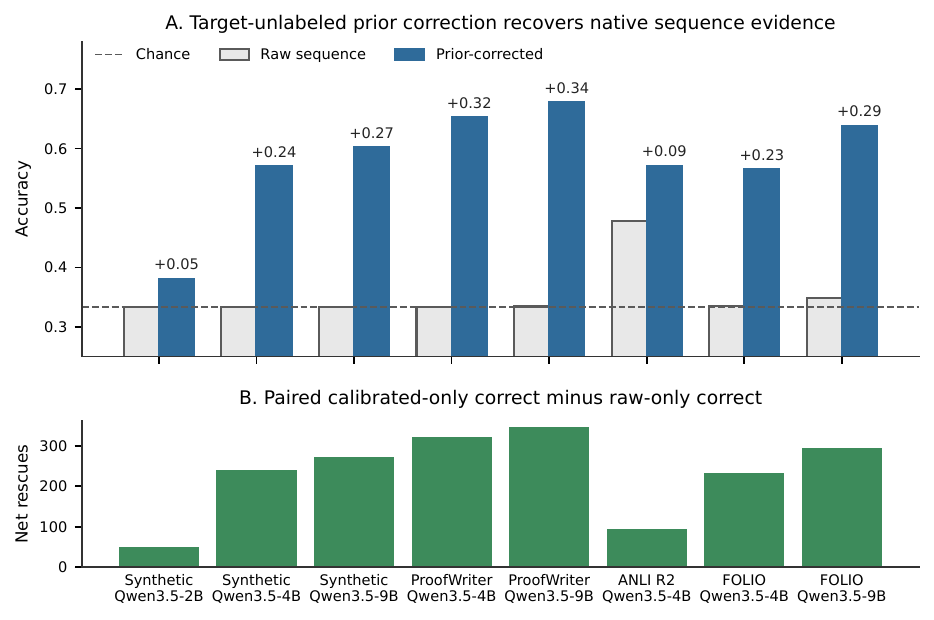}
\caption{\textbf{Cross-task recovery from collapsed sequence scoring.}
\textbf{(A)}~Raw sequence accuracy (gray) versus corrected
accuracy (blue); dashed line is chance ($1/3$). \textbf{(B)}~Paired net gain,
defined as corrected-only correct minus raw-only correct. Bars use 1,000
unlabeled ID score rows for Synthetic, ProofWriter, and ANLI, and 203 validation
rows for FOLIO; every bar is evaluated on a disjoint held-out split. The
25-example result belongs only to the separate sample-efficiency analysis.}
\label{fig:native_recovery}
\end{figure}

Table~\ref{tab:native_prior_rescue_summary} and Figure~\ref{fig:native_recovery} confirm this prediction. Fitting offsets on 1,000 unlabeled ID score rows raises \modelname{Qwen3.5-4B/9B} on the synthetic lexical split from chance to $0.570$ and $0.602$, respectively, and to $0.653$/$0.678$ on ProofWriter. Every configuration in Table~\ref{tab:native_prior_rescue_summary} reaches statistical significance ($p\le 0.035$). Even on the adversarial ANLI task, \modelname{Qwen3.5-4B} improves from $0.478$ to $0.571$.

\looseness=-1 The parallel gain in macro-F1 shows that the correction is not defaulting to a single frequent class. Before correction, \modelname{Qwen3.5-9B} predicts \unknown~for 999 of 1,000 lexical-OOD examples; after the thresholds are rebalanced---without a single target label---the relative score differences reorder the candidates correctly. Raw sequence scores thus retain per-example logic even when the argmax is dominated by label-level offsets.

\begin{table*}[t]
\centering
\small
\resizebox{\textwidth}{!}{%
% Auto-generated from experiments/revision_summaries/native_prior_rescue_summary.csv.
% Regenerate with: bash scripts/run_native_prior_rescue_summary.sh
\begin{tabular}{llcccccc}
\toprule
Task & Model & $N$ & Raw & Cal. & Paired Gain & Macro-F1 & $p$ \\
\midrule
synthetic lexical & \textit{Qwen3.5-2B} & 1000 & 0.333 & 0.381 & 272-224=+48 & 0.371 & $p=0.035$ \\
 & \textit{Qwen3.5-4B} & 1000 & 0.333 & 0.570 & 351-114=+237 & 0.572 & $p<.001$ \\
 & \textit{Qwen3.5-9B} & 1000 & 0.333 & 0.602 & 371-102=+269 & 0.591 & $p<.001$ \\
\midrule
ProofWriter & \textit{Qwen3.5-4B} & 1000 & 0.333 & 0.653 & 478-158=+320 & 0.650 & $p<.001$ \\
 & \textit{Qwen3.5-9B} & 1000 & 0.334 & 0.678 & 474-130=+344 & 0.678 & $p<.001$ \\
\midrule
ANLI R2 & \textit{Qwen3.5-4B} & 1000 & 0.478 & 0.571 & 149-56=+93 & 0.573 & $p<.001$ \\
\midrule
FOLIO diagnostic & \textit{Qwen3.5-4B} & 1001 & 0.335 & 0.565 & 407-176=+231 & 0.553 & $p<.001$ \\
 & \textit{Qwen3.5-9B} & 1001 & 0.348 & 0.638 & 437-146=+291 & 0.632 & $p<.001$ \\
\bottomrule
\end{tabular}%
}
\caption{\textbf{Sequence scores preserve answer evidence after label-bias correction.}
Additive per-label offsets are fit on unlabeled in-domain score distributions
and frozen before held-out evaluation. Per-example gain reports examples the
corrected rule gets right that the raw rule gets wrong, minus the reverse.}
\label{tab:native_prior_rescue_summary}
\end{table*}

\subsection{Sample efficiency}
\label{sec:sample-efficiency}

If the offset correction addresses a low-dimensional threshold bias rather than learning a new task mapping, it should need very few examples to fit.

Appendix~Table~\ref{tab:app_sample_efficiency} reports offset fits with as few as 25 unlabeled examples. Across all 30 random subsamples at the 25-example budget, \modelname{Qwen3.5-2B/4B/9B} show positive accuracy gains on the synthetic lexical split; on ProofWriter, the 25-example fit yields improvements of $+0.216$ and $+0.277$ for \modelname{Qwen3.5-4B/9B}. Moving from 25 to 1,000 fit examples changes held-out accuracy by at most $0.022$ on any task. This rapid saturation is consistent with fitting only two bias parameters: the representations are already separable, merely shifted relative to the output geometry, and the correction recenters them without retraining.

Table~\ref{tab:sample_efficiency_main} exposes the two budget endpoints behind this result. Recovery is already positive at 25 examples for every collapsed Qwen3.5 row, remains positive in all 30 subsamples, and changes only modestly by the full budget. The near-ceiling ANLI-9B row is the informative exception: additional unlabeled data cannot create headroom when the raw score is already strong. The complete sweep appears in the appendix.

\begin{table*}[t]
\centering
\scriptsize
\setlength{\tabcolsep}{4pt}
\begin{tabular}{llrrrr}
\toprule
Task & Model & Raw & 25 ID & 1000 ID & Pos@25 \\
\midrule
Synthetic lexical & \modelname{Qwen3.5-4B} & .333 & .549$\pm$.026 & .570 & 30/30 \\
Synthetic lexical & \modelname{Qwen3.5-9B} & .333 & .580$\pm$.024 & .602 & 30/30 \\
ProofWriter & \modelname{Qwen3.5-4B} & .333 & .641$\pm$.023 & .644 & 30/30 \\
ProofWriter & \modelname{Qwen3.5-9B} & .334 & .675$\pm$.018 & .677 & 30/30 \\
ANLI R2 & \modelname{Qwen3.5-4B} & .445 & .559$\pm$.023 & .574 & 30/30 \\
ANLI R2 & \modelname{Qwen3.5-9B} & .638 & .620$\pm$.019 & .638 & 6/30 \\
\bottomrule
\end{tabular}
\caption{\textbf{Selected sample-efficiency results.} Offsets are fit on
unlabeled in-domain score rows and evaluated on a disjoint held-out split. The
25-ID column reports mean calibrated accuracy over 30 deterministic subsamples
(standard deviation shown); 1000-ID is the full-budget result; ``Pos@25''
counts subsamples with a positive gain. Complete rows and intermediate budgets
appear in Appendix~\ref{tab:app_sample_efficiency}.}
\label{tab:sample_efficiency_main}
\end{table*}

\subsection{Shortcut and label-count controls}
\label{sec:controls}

To rule out lexical shortcuts and label-distribution matching as explanations, we stress-test the corrected decisions against two controls.

\looseness=-1 As described in Section~\ref{sec:method}, we first isolate examples that a bag-of-words (TF-IDF) classifier fails on, removing instances solvable by premise--hypothesis overlap. On these ProofWriter TF-IDF-missed subsets, the 25-example correction maintains \modelname{Qwen3.5-4B/9B} accuracies of $0.622$ and $0.643$. Second, we ask whether the gains reflect a favorable label histogram by comparing them with count-preserving permutation baselines: on balanced evaluation sets, a purely distributional fix would hover at chance ($1/3$). The corrected \modelname{Qwen3.5-4B/9B} predictions instead produce permutation gaps of $+0.287$ and $+0.305$, indicating that the accuracy reflects per-example alignment. Table~\ref{tab:control_summary} aggregates these controls; neither the collapse nor the recovery is explained away by surface heuristics.

Two further audits bound the roles of the prior and the prompt surface. Perturbing the target prior around the uniform operating point leaves worst-case gains positive for \modelname{Qwen3.5-4B/9B} on synthetic lexical ($+0.193$/$+0.242$) and ProofWriter ($+0.281$/$+0.294$), whereas \modelname{Pythia-12B} degrades to $-0.021$ (Tables~\ref{tab:app_synth_sensitivity} and~\ref{tab:app_proofwriter_native_sensitivity}). The collapse is also not tied to the answer surface: holding the verbalizers to equal token counts leaves \modelname{Qwen3.5-9B} string scoring at chance, and across 20 prompt/verbalizer variants the collapse persists in $87$--$96\%$ of configurations (Appendix~\ref{app:surface}). Finally, regenerating the evaluation splits with three independent generator seeds leaves the \modelname{Qwen3.5-9B} recovery intact (minimum $\Delta$ $+0.230$; observed-minus-null gap $+0.252$; Table~\ref{tab:app_multiseed_calibration}), so the effect is a model property rather than a dataset artifact.

The detailed TF-IDF-missed results in Table~\ref{tab:bow_control_main} make the strongest control concrete. On the hard ProofWriter slice, the 25-example correction reaches $0.622$/$0.643$ for \modelname{Qwen3.5-4B/9B}, while the observed-minus-null gaps are $+0.287$/$+0.305$ at the full budget. These margins remain well above the count-preserving null even after removing examples that a bag-of-words classifier can solve.

\begin{table*}[t]
\centering
\scriptsize
\setlength{\tabcolsep}{4pt}
\begin{tabular}{llrrrrr}
\toprule
Task & Model & $N_{\mathrm{TFIDF\mbox{-}miss}}$ & Raw & 25 ID & 1000 ID & Null gap (full) \\
\midrule
Synthetic lexical & \modelname{Qwen3.5-4B} & 645 & .468 & .558$\pm$.018 & .569 & +.187 \\
Synthetic lexical & \modelname{Qwen3.5-9B} & 645 & .468 & .581$\pm$.030 & .594 & +.212 \\
ProofWriter & \modelname{Qwen3.5-4B} & 576 & .366 & .622$\pm$.024 & .622 & +.287 \\
ProofWriter & \modelname{Qwen3.5-9B} & 576 & .366 & .643$\pm$.024 & .651 & +.305 \\
ANLI R2 & \modelname{Qwen3.5-4B} & 660 & .377 & .557$\pm$.038 & .583 & +.226 \\
\bottomrule
\end{tabular}
\caption{\textbf{Selected TF-IDF-missed recovery results.} Evaluation rows
are restricted to instances missed by a bag-of-words classifier. The 25-ID and
1000-ID columns report calibrated accuracy for offsets fit with the indicated
unlabeled budgets; ``Null gap (full)'' is the full-budget
observed-minus-count-preserving-permutation gap. Complete model and slice rows appear
in Appendix~\ref{tab:app_native_bow_sample}.}
\label{tab:bow_control_main}
\end{table*}

\begin{table}[t]
\centering
\scriptsize
\setlength{\tabcolsep}{1.6pt}
\renewcommand{\arraystretch}{0.93}
\begin{tabular}{@{}L{0.22\columnwidth}L{0.29\columnwidth}L{0.41\columnwidth}@{}}
\toprule
Control & Alternative explanation & Main outcome \\
\midrule
TF-IDF-missed examples &
Lexical shortcuts. &
\modelname{Qwen3.5-4B/9B} stay positive on synthetic and ProofWriter hard examples; \modelname{Qwen3.5-4B}
also on ANLI R2. \\
Label-count permutation baselines &
Label-distribution matching only. &
Main \modelname{Qwen3.5} rows exceed their permutation baselines; multiseed \modelname{Qwen3.5-9B} lexical-OOD gap is $+0.205$. \\
Prior perturbations &
Sensitive to the assumed prior. &
Worst-case gains remain positive on synthetic ($+0.193/+0.242$) and ProofWriter
($+0.281/+0.294$). \\
25-example fits &
Requires large calibration set. &
Collapsed rows recover from 25 unlabeled examples on synthetic, ProofWriter, and
ANLI R2. \\
Surface/model scope &
Answer-string artifact or universal effect. &
20 prompt/verbalizer variants and equal-token ABC preserve collapse; \modelname{Pythia} and
high-raw \modelname{Qwen3.5} rows delimit scope. \\
\bottomrule
\end{tabular}
\caption{\textbf{Control summary.} Corrected decisions are checked against
lexical filters, label-count permutation baselines, prior perturbations,
small-sample fits, surface changes, and model-scope controls.}
\label{tab:control_summary}
\end{table}

\begin{figure}[!htb]
\centering
\includegraphics[width=0.76\columnwidth]{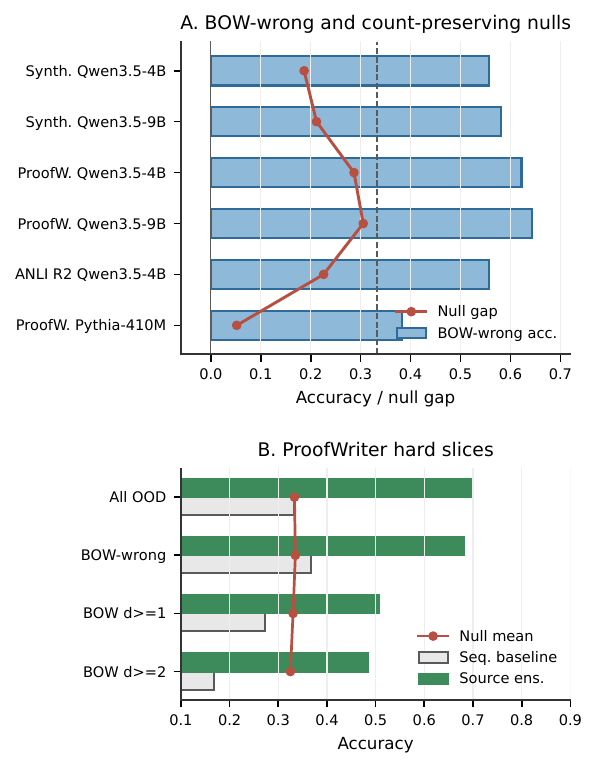}
\caption{\textbf{Shortcut and label-count controls for TF-IDF-missed examples.}
\textbf{(A)}~Bars: calibrated accuracy on examples a TF-IDF classifier misses;
red dots: gap above label-count permutation nulls.
\modelname{Qwen3.5} rows exceed both chance and nulls; \modelname{Pythia} does
not. \textbf{(B)}~ProofWriter hard slices: source ensemble (green) exceeds the
sequence baseline (gray) and null means (red) across depth strata.}
\label{fig:robustness}
\end{figure}

\subsection{Complementarity with source-task transfer}
\label{sec:hybrid}

\looseness=-1 Finally, if corrected sequence scores capture genuine target-domain evidence, they should complement representations learned through cross-task transfer: fine-tuned adapters rewrite reasoning pathways, while native sequence scores rely on pre-trained routing that may preserve complementary structure.

We test this on ProofWriter with a source-task ensemble (\modelname{Qwen3.5-4B}, \modelname{Qwen3.5-9B}, and \modelname{Llama-3.1-8B} adapters) fixed before seeing any target labels. The source-only ensemble achieves $0.700$ accuracy on ProofWriter OOD. Fusing the target-domain corrected sequence scores---still without target labels---raises the hybrid ensemble to $0.720$, a paired net gain of 20 examples over the baseline. The margin is modest and we do not attach a significance claim to it, but the direction is consistent across the hard-slice audits: on the TF-IDF-missed slice the hybrid improves from $0.682$ to $0.707$, and on the deepest proof strata ($\ge\!2$ hops) from $0.483$ to $0.557$ (Appendix~\ref{app:proofwriter}). This suggests that, once the scoring bottleneck is removed, native sequence scores carry additive evidence that source-task transfer alone does not capture.

\begin{table}[t]
\centering
\scriptsize
\setlength{\tabcolsep}{2.5pt}
\begin{tabular}{@{}lrrrr@{}}
\toprule
Model & Raw & Corrected & $\Delta$ & Permutation test \\
\midrule
\modelname{OLMo-2-1B} & .362 & .566 & +.204 & $p<.001$ \\
\modelname{Llama-3.1-8B} & .333 & .477 & +.144 & $p<.001$ \\
\modelname{Pythia-410M} & .328 & .353 & +.025 & $p=.103$ \\
\modelname{Pythia-12B} & .343 & .335 & $-.008$ & $p=.462$ \\
\bottomrule
\end{tabular}
\caption{\textbf{Cross-family ProofWriter comparison.} Offsets are fitted on
1,000 unlabeled ID score rows and frozen for a disjoint 1,000-example OOD
split. The OLMo and Llama gains exceed count-preserving permutation nulls;
the two Pythia changes are not statistically reliable and serve as scope
boundaries.}
\label{tab:cross_family_proofwriter}
\end{table}

\subsection{Boundary conditions: when recovery fails}
\label{sec:boundaries}

The correction is not expected to help universally, and its failures are informative. Three boundary cases recur across our tables. First, \emph{near-ceiling rows}: the high-baseline \modelname{Qwen3.5-9B} ANLI configuration (raw accuracy $0.638$) leaves little recoverable structure, and the offset yields no gain ($\Delta=0.000$ at the 1,000-example fit, positive in only 6 of 30 subsampled 25-example fits). Second, \emph{models without a collapsed ranking}: \modelname{Pythia-410M} and \modelname{Pythia-12B} show no reliable recovery on ProofWriter ($p=.103$ and $p=.462$; Table~\ref{tab:cross_family_proofwriter}); no hidden ranking is exposed for the offset to rescue. Third, \emph{prior misspecification}: worst-case prior perturbations shrink the gains but keep them positive on the synthetic and ProofWriter splits (Table~\ref{tab:control_summary}), so a wrong prior degrades the correction gradually rather than inverting it. These cases sharpen the scope of our claim: the correction recovers per-example evidence only when probes indicate that the answer is encoded \emph{and} raw predictions are collapsed or strongly biased, rather than merely inaccurate.

\section{Discussion}
\label{sec:discussion}

\paragraph{The conjunctive nature of evidence.}
Evaluating reasoning capacity through raw likelihood scoring underestimates a model's internal representations, but this evidence must be read conjunctively. High probe accuracy alone shows that an answer is linearly decodable from hidden states, not that the model's native scoring uses it; conversely, positive per-example gains from calibration are meaningful only if they exceed the label-count permutation baseline. Under this reading, the \modelname{Pythia} checkpoints fail the permutation-null test and the high-baseline \modelname{Qwen3.5-9B} ANLI row shows no recoverable gain, so we treat the offset as a descriptive diagnostic for collapsed distributions rather than a universal causal fix for models that lack the reasoning skill.

\paragraph{Mechanistic origins of the scoring collapse.}
The additive decomposition $s_y(x)=\tilde{s}_y(x)+\beta_y$ is effective because task-relevant variation survives in the relative score differences, even when a large label-dependent offset dominates the raw argmax; altering no within-label ranking, the two-parameter correction cannot ``teach'' new reasoning paths---it removes a shared threshold shift so that existing logic can surface. Why does the native output matrix fail to expose this structure? The Base-model results (Table~\ref{tab:qwen_base_position_matched}) show that the collapse precedes instruction tuning; the bias plausibly originates in pretraining frequency priors over structural tokens, which instruction tuning and safety alignment then amplify toward neutral or uncertainty markers (e.g., \unknown). The task-relevant signal then behaves as a small perturbation $\tilde{s}_y(x)$ on top of a much larger structural bias $\beta_y$.

\paragraph{Why recovery is partial.}
\looseness=-1 The correction closes only part of the probe--sequence gap: on the controlled-logic lexical-OOD split, corrected accuracy reaches $0.57$--$0.60$ where answer-slot probes reach $0.77$--$0.83$. This is expected under the additive account: the fitted offsets remove only the shared label-level component $\beta_y$ and cannot repair per-example distortions such as surface-form competition between candidate strings \cite{holtzman2021surface} or context-dependent misalignment of the unembedding geometry. The residual gap bounds what a global, target-label-free correction can expose and marks per-example scoring distortions as the next diagnostic target.

\paragraph{Connections to chain-of-thought prompting.}
The scoring bottleneck we identify offers a mechanistic motivation for chain-of-thought (CoT) prompting \cite{wei2022chain}: mapping a multi-hop deduction into a single classification token concentrates the representational burden on one unembedding step, where we observe the collapse, while CoT distributes the deduction across intermediate tokens and hidden states. The connection is speculative, but it predicts that the collapse should shrink as evaluation moves from single-token to free-form answers.

\paragraph{Implications for evaluation practice.}
These results argue for a change in how zero-shot failures are interpreted. Before concluding that a model lacks a reasoning capability because its accuracy is near chance, it is worth inspecting the prediction histogram: if the predictions collapse onto a single label, a lightweight, target-label-free offset fitted on a small unlabeled set can reveal whether the capability is absent or only masked by output-layer artifacts. Leaderboard rankings may currently penalize some models not for poor reasoning but for poor default calibration.

\section{Conclusion}

Our results show that a wrong prediction need not mean that the underlying answer evidence is absent. Across controlled logic, ProofWriter, ANLI, and FOLIO, a low-capacity, target-label-free offset exposes held-out, example-specific structure and recovers accuracy for the larger \modelname{Qwen3.5} models, with transfer to \modelname{OLMo} and \modelname{Llama}. These gains persist on TF-IDF-missed examples and exceed count-preserving permutation baselines, while \modelname{Pythia} and near-ceiling configurations delineate clear limits. Together, the findings argue that benchmark errors should be diagnosed across both internal representations and answer scoring before they are attributed to missing reasoning capability.

\section*{Limitations}

\looseness=-1 The supplied label prior in our diagnostic is an explicit assumption. Because every evaluation split in this paper is label-balanced, the correction is tested within the regime its prior describes. Our results therefore show that collapsed scores retain per-example evidence under a known operating regime, but they do not prescribe a solution for deployments where the true label distribution is imbalanced or unknown. Identifying both the prior and the offsets from the same unlabeled score matrix without additional supervision remains an open challenge.

Second, our method is an observational diagnostic, not a causal circuit analysis. Probes, same-position label logits, and corrected scores triangulate \emph{where} representation and scoring diverge, but they do not isolate the neural circuits responsible for the bottleneck. A positive permutation gap confirms example-specific alignment beyond label counts, yet it is not proof of human-like semantic processing. Our lexical, metadata, and matched counterfactual controls reduce shortcut explanations but cannot eliminate them.

Finally, the full position-matched chain centers on the protocol-matched \modelname{Qwen3.5} sweep. While the transfer results on \modelname{OLMo} and \modelname{Llama} are promising, the negative results on \modelname{Pythia} show that the phenomenon is not universal across architectures. Extending the framework beyond forced-choice tasks to open-ended generation would require defining robust candidate-answer strings and matched permutation controls, which falls outside the scope of this work.

\section*{Ethics Statement}

This is an interpretability and evaluation study using synthetic data, public benchmarks, and open-weight model checkpoints (\modelname{Qwen3.5}, \modelname{Pythia}, \modelname{OLMo}, and \modelname{Llama}). All artifacts are used within their respective licenses for research evaluation. The study does not collect personal data or introduce a deployed decision system. The primary ethical risk is one of overclaiming: demonstrating that corrected scores recover accuracy could be misconstrued as proof that a model reasons safely or faithfully internally. To mitigate this, we report scope controls, shortcut audits, and explicit boundaries, and frame the correction as diagnostic evidence rather than a reliability guarantee.

\section*{Acknowledgements}
This work was supported in part by the Beijing Major Science and Technology
Project under Contract No.~Z251100008125054. This work was supported by the
Beijing Academy of Artificial Intelligence (BAAI). This work was supported by
Yixin Group Limited. We gratefully acknowledge their provision of the essential
computing resources required for our experiments.

\bibliography{custom}

\appendix

\section{Appendix Overview}

The appendix remains in the same ACL two-column layout as the main text. It first records data construction, model revisions, score computation, offset fitting, and uncertainty procedures; it then reports surface and verbalizer controls, expanded readout audits, controlled-logic robustness checks, ProofWriter transfer experiments, and the ANLI/FOLIO diagnostics. Within each block, read the raw scores first, then the label-free correction, and finally the TF-IDF-missed, permutation, prior-sensitivity, and model-family controls that delimit the interpretation.

\section{Data, Models, and Reproducibility}
\label{app:data}

This section details the experimental configurations, split sizes, score computations, uncertainty procedures, and checkpoint revisions. All principal held-out evaluations use at least 1,000 examples, except for the explicitly labeled FOLIO diagnostic. Controlled logic, ProofWriter, ANLI, and FOLIO use fixed three-way label sets and a uniform prior for the diagnostic offset fit; the FOLIO fit uses 203 validation examples and evaluates on a separate 1,001-example prepared split. Controlled-logic data is generated deterministically with fixed scripts and seeds, and every evaluation split is kept disjoint from the unlabeled fit set.

\paragraph{Compute budget.}
The broader artifact suite from which this paper draws required roughly 400 H100 GPU-hours; the present paper uses a selected subset of those results. Models range from 410M to 12B parameters. Once scores are cached, each offset fit is a deterministic two-dimensional grid search and adds negligible cost.

\paragraph{Offset fitting mechanics.}
In our codebase, the additive prior offsets are identified via a deterministic two-dimensional search. By fixing one label offset to zero, we rank the candidate parameter pairs by their fit to the prior, preferring lower collapse and smaller norm. Bootstrap intervals, paired statistical tests, sampled-ID repeats, and TF-IDF baselines govern the uncertainty and robustness reporting, following the pipeline: compute score $\to$ fit offsets label-free $\to$ freeze $\to$ evaluate on held-out.

\paragraph{Format-matched reruns.}
The \modelname{Qwen3.5-4B} answer-slot table relies on the pinned revision documented in the experiment manifest: formatted prompts end precisely with the literal \texttt{Answer: <answer>} suffix, with 2,000 probe-training rows across three 1,000-example held-out splits (seeds 42/123/456), left padding, and truncation at a maximum length of 1,024. Earlier values were excluded due to misaligned \texttt{no\_format\_prompt} flags between training and evaluation. The ProofWriter \modelname{Pythia} reruns similarly use pinned revisions, with 1,000 ID and 1,000 OOD rows per checkpoint. Complete provenance is recorded in the manifest and revision-summary CSVs supplied with the submission artifacts.

\paragraph{Controlled-logic dataset.}
Our controlled-logic benchmark is a balanced three-way deductive task (\texttt{true}/\texttt{false}/\texttt{unknown}). Each example contains shuffled logical facts and rules populated with nonce terms (e.g., \textit{dax}, \textit{wug}, \textit{mip}). Distractors are label-balanced: every target label is accompanied by two positive-property and two negative-property rules targeting the query, so that solving the task requires multi-hop reasoning rather than shallow lexical matching.

\paragraph{Example (depth-2, label \texttt{true}).}
\begin{quote}\small
\textbf{Facts and rules:} Ivo is a marn. Every marn is a pavo. Every joto is
bright. No wug is bright. Every pavo is a joto. Every joto is fragile. No dax
is bright. Every sarn is bright.\\
\textbf{Question:} Is Ivo bright?\\
\textbf{Answer:} true \quad (chain: marn $\to$ pavo $\to$ joto $\to$ bright)
\end{quote}

\noindent Generalization is evaluated across the following axes:
\begin{itemize}[nosep,leftmargin=1.2em]
\item \textbf{ID test} (1,000 examples): Same depth distribution (1--2 hops) and template style as the fit set.
\item \textbf{Depth OOD} (1,000): Extended reasoning chains (3--5 hops) using identical templates.
\item \textbf{Lexical OOD} (1,000): Identical depth (1--2) with aggressively paraphrased templates (e.g., ``\textit{All \{src\} things are \{dst\} things}'').
\item \textbf{Stress/counterfactual}: Alternative generator seeds and counterfactual rule inversions to check structural robustness.
\end{itemize}
These splits share a universal nonce vocabulary and entity set; only depth, template syntax, and generator seeds vary.

\appendixcolumntable{0.98}{model_manifest_summary}{\textbf{Model manifest.} Pinned checkpoint revisions and trust status for the principal \modelname{Qwen3.5}, \modelname{Pythia}, \modelname{OLMo}, and \modelname{Llama} comparisons.}{tab:app_model_manifest}

\section{Surface and Verbalizer Controls}
\label{app:surface}

Before analyzing sequence-score recovery, we first check what shallow information the prompts might leak. The surface-leakage table tests whether the controlled-logic prompts can be solved by metadata or template structure alone. The strength of certain metadata baselines motivates our reliance on TF-IDF-missed slices and permutation nulls as the primary evidence controls.

\appendixcolumntable{0.98}{surface_leakage_audit}{\textbf{Surface leakage audit.} TF-IDF and metadata baselines test whether controlled-logic prompts reveal labels through shallow features; the resulting controls motivate TF-IDF-missed and permutation checks.}{tab:app_surface_leakage}

The prompt and verbalizer controls then verify that the initial collapse is not an artifact of answer-string length or prompt wording. The collapse persists across these variants, which indicates that the correction targets an output-layer bottleneck rather than a formatting artifact.

\appendixcolumntable{0.98}{equal_token_verbalizer_audit}{\textbf{Equal-token verbalizer audit.} Holding verbalizer token length constant, the sequence-score collapse persists, ruling out answer-length artifacts.}{tab:app_equal_token}

\appendixwidetable{0.90}{prompt_verbalizer_robustness}{\textbf{Prompt and verbalizer robustness.} The collapse and its label-free recovery persist across 20 prompt/verbalizer variants, rather than depending on one surface template.}{tab:app_prompt_verbalizer}

\section{Expanded Readout Audit}
\label{app:readout}

This section expands on the readout gap diagnostics of §\ref{sec:readout-gap}; the position-matched \modelname{Qwen3.5} Base comparison appears below (Table~\ref{tab:qwen_base_position_matched}). Table~\ref{tab:app_qwen_best_wu} further shows that even when the best native readout position is selected, answer-slot probes remain well above the best label-logit directions.

\appendixwidetable{0.90}{qwen_base_position_matched}{\textbf{Position-matched \modelname{Qwen3.5} Base scoring.} Base checkpoints show the same probe--scoring divergence; even partially informative label logits collapse again when scores are aggregated over the full candidate string.}{tab:qwen_base_position_matched}

\appendixsidebyside
{0.98}{qwen_base_best_wu_readout_synthesis}
{\textbf{\modelname{Qwen3.5} Base native-readout synthesis.} Selecting the best native readout position narrows the gap only slightly; answer-slot probes remain well ahead, leaving sequence scoring as the weakest link.}
{tab:app_qwen_best_wu}
{0.98}{sequence_scoring_sensitivity}
{\textbf{Sequence scoring sensitivity.} Mean normalization (versus sum) leaves the candidate-score collapse unchanged, ruling out sequence-length normalization as an explanation.}
{tab:app_sequence_sensitivity}

Tables~\ref{tab:app_probe_selectivity} and \ref{tab:app_probe_direction} bound what the probes can tell us. Random-label fits establish a baseline for spurious decodability, so that probe success can be read as answer-related structure rather than arbitrary side-feature memorization.

\appendixcolumntable{0.98}{probe_selectivity_controls}{\textbf{Probe selectivity controls.} Randomized and auxiliary targets bound the answer-related signal that can be decoded by the probes.}
{tab:app_probe_selectivity}

\appendixcolumntable{0.98}{probe_direction_stability}{\textbf{Probe direction stability.} Transferring answer directions across splits measures probe variance under distribution shift and bounds their use as diagnostic upper limits.}{tab:app_probe_direction}

Finally, Tables~\ref{tab:app_readout_gap_sig} and~\ref{tab:app_probe_size} complete the probe audit: the first quantifies the readout gap with bootstrap intervals and paired tests, and the second measures how probe accuracy depends on labeled training-set size.

\appendixwidetable{0.92}{readout_gap_significance}{\textbf{Probe/readout gap significance.} Bootstrap intervals and paired statistical tests quantify the gaps between linear probes, native readouts, and full-string scoring.}{tab:app_readout_gap_sig}
\appendixwidetable{0.98}{probe_training_size_sensitivity}{\textbf{Probe training-size sensitivity.} We measure how probe accuracy scales with training-set size; the probes behave as stable diagnostic upper bounds whose properties are cross-checked by the zero-shot sequence tests.}{tab:app_probe_size}

\section{Controlled-Logic Parallel Results and Label-Bias Details}
\label{app:native}

This section extends §\ref{sec:sequence-recovery} and §\ref{sec:controls} across seeds, alternate score normalizations, and TF-IDF-missed subsets. The \modelname{Pythia-2.8B/12B} checkpoints serve as boundary controls throughout. Across multiple independent generators, the \modelname{Qwen3.5} models recover consistently, arguing against explanations based on score-normalization artifacts.

\appendixwidetable{0.92}{synthetic_sequence_prior_calibration}{\textbf{Controlled-logic label-bias correction.} Offsets fit on unlabeled ID scores recover held-out accuracy and macro-F1 for larger \modelname{Qwen3.5} checkpoints, while \modelname{Pythia} rows mark the boundary.}
{tab:app_synth_calibration}

\appendixwidetable{0.95}{synthetic_sequence_prior_sensitivity}{\textbf{Controlled-logic prior sensitivity.} We perturb the target prior to stress-test the offset fit. Positive margins persist for \modelname{Qwen3.5}, indicating that the recovery relies on internal rankings rather than a precisely tuned prior.}
{tab:app_synth_sensitivity}

\appendixwidetable{0.96}{native_prior_sample_efficiency_qwen_main}{\textbf{Offset sample efficiency.} Held-out recovery is already positive with 25 unlabeled fit examples and saturates quickly, as expected for a two-parameter threshold shift.}
{tab:app_sample_efficiency}

\begin{figure}[t]
\centering
\includegraphics[width=0.72\linewidth]{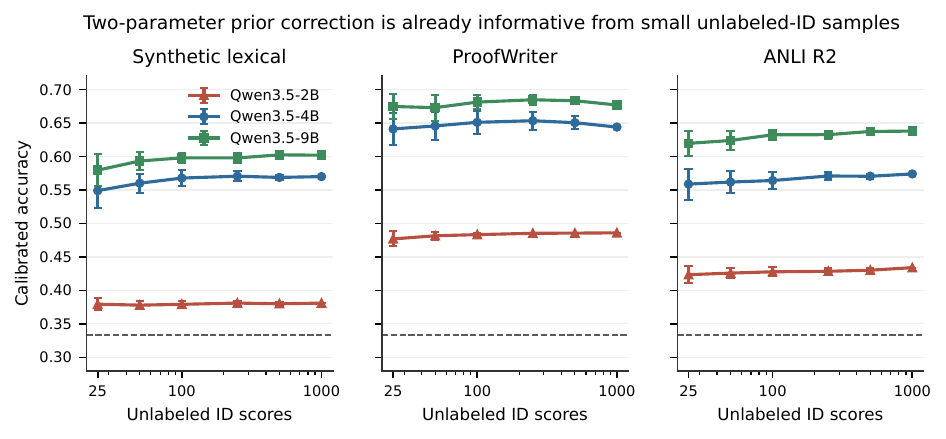}
\caption{\textbf{Offset sample efficiency.} Calibrated accuracy as the
unlabeled ID fit budget grows (log scale; error bars show seed variance). The
larger \modelname{Qwen3.5} checkpoints are already above chance at 25 examples;
the near-ceiling ANLI-9B row is a scope control.}
\label{fig:sample_efficiency}
\end{figure}

\paragraph{Normalization and permutation nulls.}
We verify that the findings are not artifacts of length normalization or label-distribution matching. Table~\ref{tab:app_synth_mean} replicates the phenomenon with mean-normalized scores, and Table~\ref{tab:app_synth_shift_null} evaluates against count-preserving permutation baselines, supporting the claim of instance-specific alignment.

\appendixwidetable{0.88}{synthetic_sequence_prior_mean_robustness}{\textbf{Mean-normalized controlled-logic robustness.} The collapse and the recovery persist under mean-normalized candidate scores, ruling out token-length artifacts.}{tab:app_synth_mean}

\appendixwidetable{0.92}{synthetic_sequence_prior_shift_permutation_null}{\textbf{Controlled-logic shift/permutation null.} Corrected predictions are compared against random permutations that share the same label counts. A positive gap indicates per-example recovery beyond histogram matching.}{tab:app_synth_shift_null}

\appendixwidetable{0.92}{synthetic_sequence_prior_multiseed_calibration}{\textbf{Independent synthetic-seed label-bias correction.} Regenerating the evaluation splits with new lexical-OOD seeds shows that the offset recovery in \modelname{Qwen3.5-9B} is a model property, not a dataset artifact.}{tab:app_multiseed_calibration}

\appendixwidetable{0.92}{synthetic_sequence_prior_multiseed_bow_sensitivity}{\textbf{Independent synthetic-seed TF-IDF-missed sensitivity.} For each seed, we recompute the TF-IDF-missed subsets; the positive margins persist, ruling out lexical-shortcut explanations.}{tab:app_multiseed_bow_sensitivity}

\appendixwidetable{0.92}{synthetic_sequence_prior_multiseed_bow_sensitivity_null}{\textbf{Independent synthetic-seed TF-IDF-missed nulls.} Seed-specific predictions must clear their label-count permutation baselines on TF-IDF-missed slices; this is the strictest combined control in our protocol.}{tab:app_multiseed_bow_null}

\paragraph{TF-IDF-missed slices summary.}
By removing instances that bag-of-words heuristics solve, these tables create an adversarial evaluation regime. Beating the permutation baseline on these slices is our strongest evidence that the prior-conditioned offset recovers example-level structure.

\appendixstackedtwotables
{0.99}{native_prior_bow_sample_efficiency}
{\textbf{TF-IDF-missed and label-count controls.} Offset corrections evaluated only on instances the TF-IDF baseline misses. The observed-minus-null gap isolates example-level structure from lexical cues.}
{tab:app_native_bow_sample}
{0.95}{synthetic_sequence_prior_multiseed_bow_stratified}
{\textbf{Independent synthetic-seed TF-IDF-missed check.} The shortcut-controlled recovery generalizes across generator seeds (52, 62, and 72).}
{tab:app_synth_multiseed_bow_main}

\section{ProofWriter Sequence Scores and Source-Task Evidence}
\label{app:proofwriter}

This section carries the synthetic methodology over to the natural-language deduction of ProofWriter. The first set of tables checks that the offset correction is not a byproduct of sequence length (mean-normalization check) or minor prior mismatch; the later tables assess the synergy between target-domain sequence scores and fine-tuned source adapters.

\appendixsidebyside
{0.98}{proofwriter_native_prior_mean_robustness}
{\textbf{ProofWriter sequence-score mean robustness.} The offset correction on mean-normalized scores replicates the recovery observed on the synthetic tasks.}
{tab:app_proofwriter_native_mean}
{0.98}{proofwriter_native_prior_sensitivity}
{\textbf{ProofWriter label-bias sensitivity.} Prior perturbations define the range in which the \modelname{Qwen3.5} models retain positive gains.}
{tab:app_proofwriter_native_sensitivity}

As on the synthetic tasks, any recovery on ProofWriter must also clear the label-count permutation null: shuffling the corrected predictions while preserving the label histogram tests whether the gains are example-specific.

\appendixstackedtwotables
{0.98}{proofwriter_native_prior_permutation_null}
{\textbf{ProofWriter label-count permutation.} The corrected accuracy exceeds what a label-matched random permutation achieves.}
{tab:app_proofwriter_native_null}
{0.92}{proofwriter_source_native_hybrid_ensemble}
{\textbf{ProofWriter source-transfer plus sequence-score hybrid.} Source adapters are frozen before any target labels are seen. Fusing them with the target sequence scores yields additive gains.}
{tab:app_proofwriter_source_ensemble}

Further ablations isolate the components behind the hybrid gain, mapping weight sensitivity and stability under small data budgets.

\appendixstackedtwotables
{0.82}{proofwriter_source_ensemble_hard_slices}
{\textbf{ProofWriter hard-slice audit for the source-task ensemble.} The sequence-augmented hybrid exceeds the frozen baseline and the permutation nulls across deeper proof strata and TF-IDF-missed variants.}
{tab:app_proofwriter_hard_slices}
{0.86}{proofwriter_source_native_hybrid_ablation}
{\textbf{ProofWriter source/sequence-score hybrid ablation.} Omitting individual sequence-score families shows that the hybrid gain draws on complementary evidence rather than a single architecture.}
{tab:app_proofwriter_hybrid_ablation}

\appendixwidetable{0.79}{proofwriter_source_native_hybrid_weight_sensitivity}{\textbf{ProofWriter source/sequence-score weight sensitivity.} Varying the sequence-score fusion weights shows a wide plateau of improvement; the complementarity does not depend on fragile tuning.}{tab:app_proofwriter_weight}

\appendixstackedtwotables
{0.79}{proofwriter_source_native_hybrid_sample_efficiency}
{\textbf{ProofWriter source/sequence-score sample efficiency.} Fits with small unlabeled budgets remain stable across hard slices, mirroring the sample efficiency seen on the synthetic tasks.}
{tab:app_proofwriter_hybrid_sample}
{0.52}{proofwriter_source_native_hybrid_prior_sensitivity}
{\textbf{ProofWriter source/sequence-score prior sensitivity.} The operating range of the label-free hybrid ensemble under prior perturbations.}
{tab:app_proofwriter_hybrid_prior}

\appendixwidetable{0.84}{proofwriter_source_ensemble_member_ablation}{\textbf{ProofWriter source-ensemble member ablation.} Multiple member subsets independently retain positive target gains, so the ensemble's contribution is not carried by a single member.}{tab:app_proofwriter_source_ablation}

\section{ANLI and FOLIO Diagnostics}
\label{app:external}

To map the generalization boundaries of the scoring bottleneck, we extend the protocol to adversarial NLI (ANLI) and first-order logic in natural language (FOLIO). ANLI removes the rigid template structure of the synthetic tasks, while FOLIO tests the diagnostic under a sparse fit set (203 examples).

\paragraph{ANLI context.}
ANLI is adversarial and far less homogeneous than controlled logic. Persistence here---under permutation baselines and small-budget fits---indicates that the offset correction extends to less templated natural-language inference.

\appendixstackedtwotables
{0.86}{anli_native_adapter_hybrid_sample_efficiency}
{\textbf{ANLI sequence-score/adapter hybrid sample efficiency.} Evaluated on R2/OOD, the hybrid remains stable at very small fitting budgets.}
{tab:app_anli_sample}
{0.86}{anli_native_adapter_hybrid_prior_sensitivity}
{\textbf{ANLI sequence-score/adapter hybrid prior sensitivity.} The TF-IDF-missed gaps persist under prior perturbations, so the gains are not an artifact of a tuned class prior.}
{tab:app_anli_prior}

\appendixwidetable{0.70}{anli_sequence_prior_permutation_null}{\textbf{ANLI sequence-score label-count permutation.} The positive \modelname{Qwen3.5} recovery gap exceeds matched label-count permutations.}{tab:app_anli_null}

\paragraph{FOLIO context.}
FOLIO is the most constrained diagnostic: offsets are fitted on 203 validation examples and evaluated on 1,001 prepared training examples.

\appendixwidetable{0.70}{folio_native_prior_calibration}{\textbf{FOLIO diagnostic label-bias correction.} This configuration is a transfer diagnostic under data sparsity, testing the offset correction outside standard benchmark splits.}{tab:app_folio_calibration}

\appendixwidetable{0.64}{folio_native_prior_sample_efficiency}{\textbf{FOLIO diagnostic sample efficiency.} Within the 203-example fit set, the \modelname{Qwen3.5} models retain positive gains at fractional subsamples.}{tab:app_folio_sample}

\appendixinlinecolumntable{0.98}{folio_native_prior_permutation_null}{\textbf{FOLIO diagnostic label-count permutation.} The 203-example fit clears label-histogram permutation nulls.}{tab:app_folio_null}

\appendixinlinecolumntable{0.98}{folio_native_prior_sensitivity}{\textbf{FOLIO diagnostic prior sensitivity.} Local shifts in the target prior do not break the recovery from the sparse validation fit.}{tab:app_folio_sensitivity}

\appendixinlinecolumntable{0.98}{folio_native_prior_bow_stratified}{\textbf{FOLIO diagnostic TF-IDF-missed check.} The limited-data diagnostic retains per-example gains on TF-IDF-missed slices of the first-order logic task.}{tab:app_folio_bow}

\section{Auxiliary Evidence and Scope Controls}
\label{app:scope}

This final section marks the experimental boundaries. The auxiliary ablations and intervention tests localize candidate mechanisms, while prompt grids record surface sensitivity. Large adapter sweeps map the engineering search space; the main text reports only selected label-free summaries. Full-split sequence-score recovery remains the primary evidence, with per-example gains and permutation nulls separating it from prompt or label-prior effects.

Read in order---setup controls, readout audits, permutation checks, and scope controls---the appendix defines the boundary conditions of the output-scoring bottleneck.

The auxiliary ablations are diagnostic rather than additional claims about benchmark performance. Component patching and prompt grids identify which parts of the scoring path are sensitive to surface form, while the label-free sequence-score tables evaluate the complete held-out split. The adapter sweeps serve a complementary purpose: they show how much of the target-domain signal is already present in a transferred representation before native scores are added.

The model-family controls are equally important for interpretation. Positive permutation gaps for the larger Qwen3.5, OLMo, and Llama configurations coexist with null or negative gaps for the Pythia checkpoints and with near-ceiling rows where no headroom remains. Taken together, these boundary cases rule out a universal calibration effect and support the narrower conclusion that a global offset can expose example-specific evidence only when the raw ranking is collapsed but the underlying representations remain informative.
\end{document}